\documentclass[runningheads]{llncs}
\usepackage{graphicx}

\usepackage{amsmath}
\usepackage{amssymb}

\usepackage{bbding}
\usepackage{pifont}
\usepackage{wasysym}
\usepackage{utfsym}
\usepackage{fontawesome}
\usepackage{url}
\usepackage{subcaption}
\usepackage{mathtools}
\usepackage{color}

\def\ie{\textit{i.e.}}

\usepackage{hyperref}
\hypersetup{hidelinks,
	colorlinks=true,
	allcolors=black,
	pdfstartview=Fit,
	breaklinks=true}

\begin{document}
%

\title{Free Lunch for Surgical Video Understanding by Distilling Self-Supervisions}

%
\author{
Xinpeng Ding\inst{1} 
\and
Ziwei Liu\inst{2} 
\and
Xiaomeng Li\inst{1, 3}
\thanks{Corresponding Authors: {\tt\small eexmli@ust.hk}}}

\authorrunning{Xinpeng Ding \and
Ziwei Liu \and
Xiaomeng Li$^\star$}
%
\institute{
The Hong Kong University of Science and Technology \and
S-Lab, Nanyang Technological University \and
The Hong Kong University of Science and Technology Shenzhen Research Institute
}
%

%
\maketitle              
\begin{abstract}
Self-supervised learning has witnessed great progress in vision and NLP; recently, it also attracted much attention to various medical imaging modalities such as X-ray, CT, and MRI. Existing methods mostly focus on building new pretext self-supervision tasks such as reconstruction, orientation, and masking identification according to the properties of medical images. However, the publicly available self-supervision models are not fully exploited.
In this paper, we present a powerful yet efficient self-supervision framework for surgical video understanding. Our key insight is to distill knowledge from \emph{publicly available models trained on large generic datasets}\footnote{For example, the released models trained on ImageNet by MoCo v2: \textcolor{blue}{\url{https://github.com/facebookresearch/moco}}} to facilitate the self-supervised learning of surgical videos.
To this end, we first introduce a semantic-preserving training scheme to obtain our teacher model, which not only contains semantics from the publicly available models, but also can produce accurate knowledge for surgical data.
Besides training with only contrastive learning, we also introduce a distillation objective to transfer the rich learned information from the teacher model to self-supervised learning on surgical data.
Extensive experiments on two surgical phase recognition benchmarks show that our framework can significantly improve the performance of existing self-supervised learning methods. Notably, our framework demonstrates a compelling advantage under a low-data regime.  
Our code is available at \textcolor{blue}{\url{https://github.com/xmed-lab/DistillingSelf}}.

\if 1 
Annotating surgical videos requires professional surgeons, which is time-consuming and cost.
To improve the performance with the limited labeled surgical data, the common approaches use the model pre-trained on the large dataset,~\emph{i.e.}, ImageNet, and then fine-tune it on the target surgical datasets.
Although great success has achieved, there is a large domain gap between ImageNet and surgical videos.

In this paper, we aims to explore the effect of self-supervised pre-training on surgical videos for surgical phase recognition.
An intuitive approach is to leverage existing contrastive learning methods to train the model on surgical videos to mitigate this domain gap.
However, this native way ignores the powerful representation ability learned from ImageNet.
\fi

\end{abstract}

\section{Introduction}
Generally, training deep neural networks requires a large amount of labeled data to achieve great performance.
However, obtaining the annotation for surgical videos is expensive as it requires professional knowledge from surgeons.
To address this problem, self-supervised learning, \ie, training the model on a large dataset (\emph{e.g.}, ImageNet~\cite{russakovsky2015imagenet}) without annotations and fine-tuning the classifier on the target datasets, has achieved great success for image recognition and video understanding~\cite{he2020momentum,chen2020improved,caron2020unsupervised,pan2021videomoco}.
This paper focuses on designing self-supervised learning methods for surgical video understanding with a downstream task - surgical phase recognition, which aims to predict what phase is occurring for each frame in a video~\cite{blum2010modeling,zappella2013surgical,jin2017sv,jin2020multi,jin2021temporal,ding2021exploiting,gao2021trans,wang2022less,twinanda2016endonet,yi2021not}.

\begin{figure}[t]%
\centering
    \includegraphics[width=0.9 \columnwidth, height=0.12\textheight]{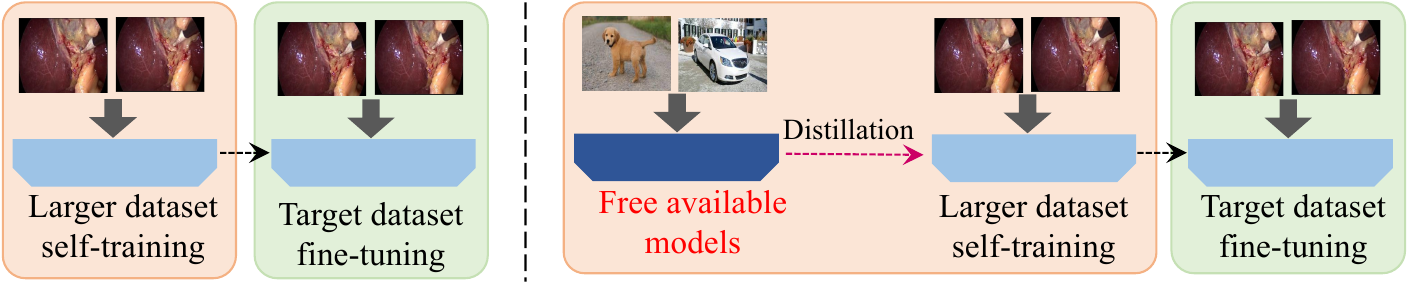}%
    \caption{
    %
    \textbf{Left:} Existing self-supervised learning approaches aim to design different pretext tasks on a larger dataset to improve the representation ability. 
    \textbf{Right:} We proposed a novel distilled contrastive learning to transfer the powerful representation ability from the free available models to the surgical video self-training.
    }
    \label{fig:difference}%
\end{figure}

Self-supervised learning has been widely applied into various medical images, such as X-ray~\cite{zhou2021preservational}, fundus images~\cite{li2021rotation,li2020self}, CT~\cite{zhuang2019self} and MRI~\cite{zhou2021models,zhou2021preservational}.
For example, Zhuang~\emph{et al.}~\cite{zhuang2019self} developed a rubik’s cube playing pretext task to learn 3D representation for CT volumes.
Rubik’s cube+~\cite{zhu2020rubik} introduced more pretext tasks,~\emph{i.e.}, cube ordering, cube orientation and masking identification to improve the self-supervised learning performance.
Some researchers introduced reconstruction of corrupted images~\cite{chen2019self,zhou2021models} or triplet loss~\cite{xie2020instance} for self-supervised learning on nuclei images.
Recently, motivated by the great success of contrastive learning in computer vision~\cite{he2020momentum,chen2020improved,tian2020contrastive,caron2020unsupervised,chen2021exploring}, Zhou~\emph{et al.}~\cite{zhou2020comparing} introduced the contrastive loss to 2D radiographs.
Taleb~\emph{et al.}~\cite{taleb20203d} further improved contrastive predictive coding~\cite{van2018representation} to a 3D manner for training with 3D medical images.
To learn more detailed context, PCRL~\cite{zhou2021preservational} combined the construction loss with the contrastive one. 
However, all of these above approaches aim to design different pretext tasks to perform self-supervised learning on medical datasets, as shown in Fig.~\ref{fig:difference}~(a).

%





%

%
In this paper, we make a crucial observation that \emph{using the same backbone and self-supervised learning method, the model trained with ImageNet data can yield a comparable performance for surgical phase recognition with that trained with surgical video data (82.3\% vs 85.7\% Acc.)}; see details in Table~\ref{tab:sota}. This surprising result indicates that the model self-supervised trained with ImageNet data can learn useful semantics that benefit surgical video understanding. There are many publicly available self-supervised models, motivating us to leverage the free knowledge to facilitate the self-supervised learning of the surgical video. 

%


To this end, we propose a novel distilled self-supervised learning method for surgical videos, which distillates the free knowledge from \emph{publicly available self-supervised models (e.g., MoCo v2 trained on the ImageNet)} to improve the self-supervised learning on surgical videos, as shown in Fig.~\ref{fig:difference}~(b).
We first introduce a semantic-preserving training to train a teacher model, which retains representation ability from ImageNet while producing accurate semantics knowledge for surgical data.
%
%
Besides training the student model with the only contrastive objective, we boost the self-supervised learning on the surgical data with a distillation objective.
The proposed distillation objective forces the similarity matrix of the teacher and the student models to be consistent, which makes the student model to learn extra semantics.
%

%
%
%

%
%
%

We summarize our key contributions as follows:
\begin{itemize}
    \item 
    To best of our knowledge, we are the first to investigate the use of self-supervised training on surgical videos.
    Instead of designing a new pretext task, we provide a new insight that to transfer knowledge from large public dataset improves self-training on surgical data.
    \item We propose a semantic-preserving training to train a teacher model, which contains representation ability from ImageNet while producing accurate semantic information for the surgical data. 
    \item We propose a distillation objective to enforce the similarity matrix between the teacher model and the student model to be consistent, which makes the student model to learn extra information during self-training.
\end{itemize}

\section{Methodology}

This section is divided into three main parts.
Firstly, we introduce contrastive learning to perform self-supervised learning on surgical videos in  Section~\ref{sec:contrastive}.
Then, the semantic-preserving training will be presented in Section~\ref{sec:semantic}.
Finally, we will describe the distilled self-supervised learning, which transfers useful semantics from publicly available models to the self-training on surgical videos.

\subsection{Contrastive Learning on Surgical Videos}\label{sec:contrastive}
Given a frame $v$ sampled from the  surgical videos, we apply two different transformations to it, which can be formally as:
\begin{equation}
    x^q = t_q(v), \ \ x^k =t_k(v),
\end{equation}
where $t_q (\cdot)$ and $t_k(\cdot)$ are two transformations, sampled from the same transformation distribution.
Then, we feed $x^q$ into a query encoder $f_q(\cdot)$, followed by a projection head $f_h(\cdot)$ consists of 2-layer multilayer perceptron (MLP), to obtain the query representation $q = h_q(f_q(x^q))$.
Similarly, we can obtain $k_+ = h_k(f_k(x^k))$, where $k_+$ indicates the positive embedding for $q$.
%
%
Similar as \cite{he2020momentum,chen2020improved}, we obtain a set of encoded samples $B = \{k_i \}_{i=1}^M$ that are the keys of a dictionary, where $k_i$ is the $i$-th key in the dictionary, and $M$ is the size of the dictionary.
There is a single key in the dictionary,~\emph{i.e.}, $k_+$, that matches $q$.
%
\begin{figure}[t]%
\centering
    \includegraphics[width=1 \columnwidth]{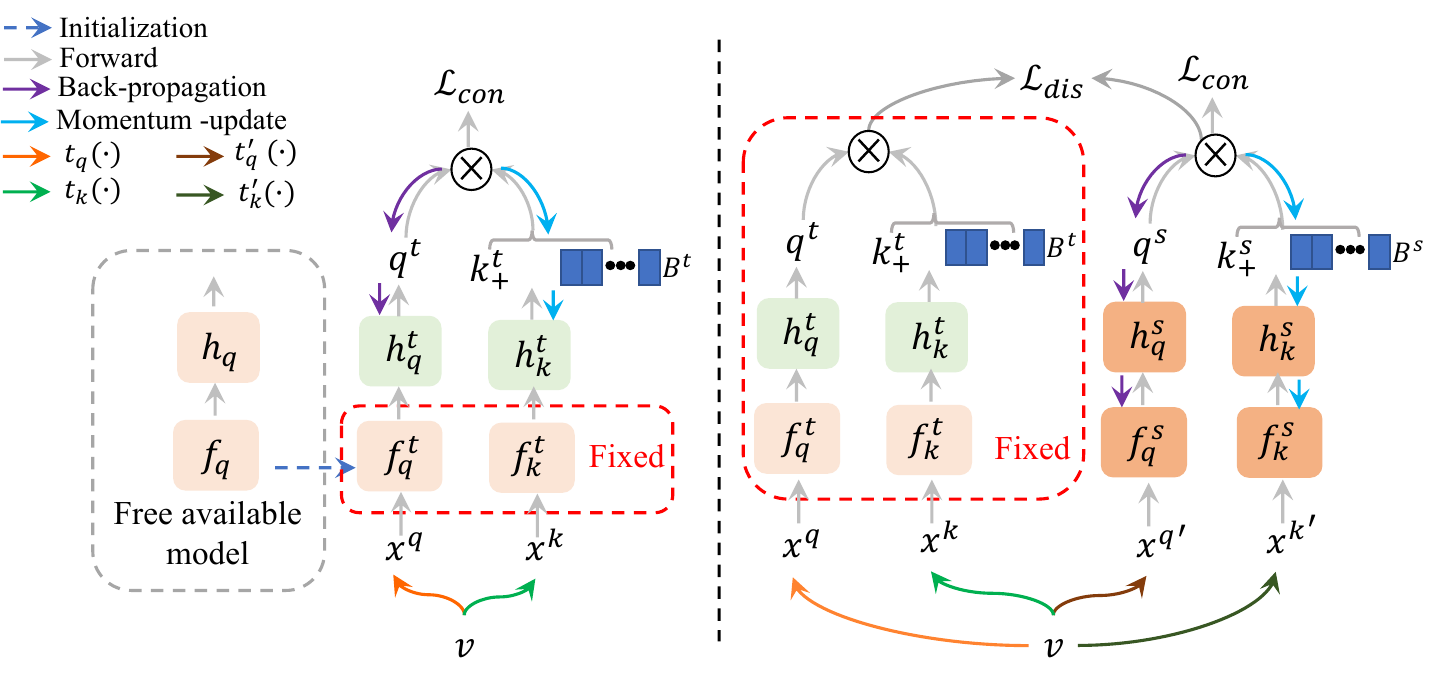}%
    \caption{
     \textbf{Left:} A semantic-preserving training is adopted to train a teacher model,~\emph{i.e.},~$f^t_q$, $f^t_k$, $h^t_q$ and $h^t_k$, from the free available model. 
     \textbf{Right:} We introduce a distilled self-supervised learning to distill the knowledge from the teacher model to the student model,~\emph{i.e.}, $f^s_q$, $f^s_k$, $h^s_q$ and $h^s_k$. 
    }
    \label{fig:framework}%
\end{figure}
In this paper, we use InfoNCE~\cite{van2018representation}, which measures the similarity via dot production, and our contrastive objective is as follows: 
\begin{equation}
    \mathcal{L}_{con}=-\sum_{i=1}^N \log \frac{\exp \left( \text{sim}(q, k_+) / \tau\right)}{ \sum_{i=1}^{M} \exp \left( \text{sim}(q, k_{i}) / \tau\right)},
    \label{E:con}
\end{equation}
where $\text{sim}(q, k_+)$ is the cosine similarity between $q$ and $k_+$, and $\tau$ is the temperature that controls the concentration level of the sample distribution~\cite{ding2021support}.
In this paper, the parameters of the query encoder are updated by back-propagation, while the key encoder is momentum-updated encoder~\cite{chen2020improved,he2020momentum}.
Let define the parameters of $f_q(\cdot)$, $h_q(\cdot)$, $f_k(\cdot)$ and $h_k(\cdot)$ as $\theta^f_q(\cdot)$, $\theta^h_q(\cdot)$, $\theta^f_k(\cdot)$ and $\theta^h_k(\cdot)$ respectively.
Then, the momentum-updating can be formulated as:
\begin{equation}
\theta^f_{\mathrm{k}} \leftarrow m \theta^f_{\mathrm{k}}+(1-m) \theta^f_{\mathrm{q}},
\end{equation}
\begin{equation}
\theta^h_{\mathrm{k}} \leftarrow m \theta^h_{\mathrm{k}}+(1-m) \theta^h_{\mathrm{q}},
\end{equation}
where $m \in [0,1)$ is a momentum coefficient, and is set to $0.999$ following~\cite{chen2020improved}.

\subsection{Semantic-Preserving Training of Free Models}\label{sec:semantic}
Our motivation is to leverage the semantic information from public free self-supervised model (\emph{i.e.},~a teacher model) trained on large datasets (\emph{e.g.}, ImageNet) to improve the self-supervised learning, (\emph{i.e.}, a student model) on surgical videos.
To this end, the teacher model should have two properties.
(1) It should retain representation ability learned from ImageNet.
Directly self-training on surgical video would make the model forget the learned knowledge from ImageNet.
(2) It should produce accurate structured knowledge for surgical videos.
Since there is a clear domain gap between ImageNet and surgical videos, transferring the noisy distilled information would degrade the performance.

To achieve these two properties, we propose the semantic-preserving training of free models on surgical video, which is shown in the left of Fig.~\ref{fig:framework}.
More specifically, we first initialize the parameters of $f_q ( \cdot)$ and $f_k ( \cdot)$ from the ImageNet pre-trained model,~\emph{i.e.}, using the weights of the free self-supervised model: Moco v2, which can be formulated as $f_q ( \cdot) \rightarrow f^t_q ( \cdot)$ and $f_k ( \cdot) \rightarrow f^t_k ( \cdot)$.
During semantic-preserving training, we fixed the parameters of the backbone,~\emph{i.e.}, $f^t_q ( \cdot)$ and $f^t_k ( \cdot)$, to maintain the prior learned semantics.
Furthermore, to produce accurate predictions for latter distillation under the large domain gap, we conduct contrastive learning on surgical video to update the parameters of the two-layer MLP,~\emph{i.e.},  $h^t_q(\cdot)$ and  $h^t_k(\cdot)$, via the contrastive objective.
In this way, the trained model would learn to use the prior knowledge from ImageNet to discriminate the surgical frames, and the predictions from it would contain rich semantic information. 
%
%
\subsection{Distilled Self-supervised Learning}\label{sec:dcl}
Unlike existing self-supervised learning approaches that focus on designing new pretext tasks, we aim to transfer the semantics knowledge from free public self-supervised models into the surgical video self-training.
As shown in the right of Fig.~\ref{fig:framework}, besides the standard contrastive learning, we also introduce a distillation objective to transfer the semantics from the teacher model (trained in Section~\ref{sec:semantic}) to the student model,~\emph{i.e.}, the model learned by self-supervised on surgical videos.
The similarity matrix (\ie$, \text{sim}(q,k_i)$) obtained in Eq.~\ref{E:con} measures the similarities between the query and its keys.
The teacher model can generate around 82.3\% Acc. for surgical phase recognition, showing that the similarity matrix contains useful semantics information that can benefit to surgical video understanding.

%

To leverage this information, our idea is to force the similarity matrix of the teacher and student model to be consistent; therefore, the teacher model can provide additional supervision for training the student model, leading to performance improvement. 
Specifically, we regard the similarity matrix of the teacher model as the soft targets to supervise the self-training of the student. 
%
To obtain the soft targets, we apply Softmax with the temperature scale $\tau$ to the similarity matrix of the teacher model.
We find that the soft targets can be formulated as:
\begin{equation}
 p_i^t =    \frac{\exp \left( sim(q^t, k^t_i) / \tau\right)}{ \sum_{i=j}^{M} \exp \left( sim(q^t, k^t_{j}) / \tau\right)},
 \label{E:sim}
\end{equation}
which is very similar to Eq.~\ref{E:con}.
Similarly, we can also obtain the similarity matrix of the student model, which can be defined as $p^s_i$.
Finally, the distillation loss is computed by the KL-divergence loss to measure the distribution of $p^t_i$ and $p^s_i$ as follows:
\begin{equation}
   \mathcal{L}_{dis} = \sum_{i=1}^M p^t_i \log \left(\frac{p^t_i}{p^s_{i}}\right),
   \label{E:dis}
\end{equation}
where $p^s_i$ is the similarity matrix of the student model, computed as in Eq.~\ref{E:sim}.
Note that the distillation loss can be regarded as a soft version of the contrastive loss (defined in Eq.~\ref{E:con}).
This soft version provides much more information per training case than hard targets~\cite{hinton2015distilling}.



Combined with the distillation loss and the contrastive loss, the overall object of the distilled self-supervised can be formulated as:
\begin{equation}
    \mathcal{L} = \mathcal{L}_{con} + \lambda \mathcal{L}_{dis},
\end{equation}
where $\lambda$ is the trade-off hyper-parameter.

\section{Experiments}

\subsection{Datasets}

\textbf{Cholec80.} The Cholec80 dataset~\cite{twinanda2016endonet} contains 80 cholecystectomy surgeries videos.
All videos are recorded as $25$ fps, and the resolution of each frame is $1920 \times 1080$ or $854 \times 480$.
We sample the videos into $5$ fps for reducing computation cost.
Following the same setting in previous works, we split the datasets into $40$ videos for training and rest $40$ videos for testing~\cite{jin2020multi,yi2021not}.

\noindent\textbf{MI2CAI16.} There are $41$ videos in the MI2CAI16~\cite{stauder2016tum} dataset, and the videos in it are recorded at $25$ fps.
Among them, $27$ videos are used for training and $14$ videos are used for test, following previous approaches~\cite{jin2020multi,yi2021not}.
%


\subsection{Implementation Details}
We use the ResNet50~\cite{he2016deep} as the backbone to extract features for each frame.
After that, following~\cite{twinanda2016endonet,yi2021not,ding2021exploiting}, a multi-stage temporal convolution (MS-TCN)~\cite{farha2019ms} is used to extracted temporal relations for frame features and predicts phase results.
Following the same evaluation protocols ~\cite{twinanda2016endonet,jin2017sv,jin2020multi,jin2021temporal}, we employ four commonly-used metrics, i.e., accuracy (AC), precision (PR), recall (RE), and Jaccard (JA) to evaluate the phase prediction accuracy.
To evaluate the self-supervised training performance, we self-train the ResNet50 backbone on surgical videos on training set of Cholec80 by different self-supervised learning approaches~\cite{chen2020improved,caron2020unsupervised,chen2021exploring,xu2021regioncl}.
Then, we linearly fine-tune the self-trained model on the training set of Cholec80 and MI2CAI16.
After that, we extract features for each frame of videos by the fine-tuned model.
Finally, based on the fixed extracted features, we use MS-TCN to predict results for surgical phase recognition.
We set $\tau$ to $0.07$ and $\lambda$ to $5$ respectively.


\subsection{Comparisons with the State-of-the-art Methods}

\begin{table}[t]
\centering
\caption{Results on Cholec80 dataset and M2CAI16 dataset. ``Scratch'' indicates training from scratch. ``IN-Sup'' and ``IN-MoCo v2'' refer to supervised learning and self-supervised learning on ImageNet, respectively.}\smallskip
\label{tab:sota}
\resizebox{1.0\columnwidth}{!}{
\begin{tabular}{c| c c c c | c c c c}
%
\hline
&\multicolumn{4}{c|}{Cholec80} & \multicolumn{4}{c}{M2CAI16} \\
\hline
   Methods & Accuracy & Precision & Recall & Jaccard & Accuracy & Precision & Recall & Jaccard\\
\hline
   Scratch & 65.0 $\pm$ 13.8 & 67.7 $\pm$ 14.3 & 53.0 $\pm$ 21.4 & 40.7 $\pm$ 18.4 &59.8 $\pm$12.9 & 65.0 $\pm$ 13.6 & 56.6 $\pm$ 29.7 & 40.0 $\pm$ 23.0\\
   \hline  
  \multicolumn{9}{c}{Models Trained on ImageNet} \\
  \hline 
   IN-Sup & 83.1 $\pm$ 10.5 & 64.7 $\pm$ 11.0 & 82.3 $\pm$ 7.6 & 77.8 $\pm$ 9.8 &
   76.0 $\pm$  6.3 & 78.8 $\pm$  5.2 &  73.8 $\pm$ 17.6 & 58.9 $\pm$ 15.0 \\
   IN-MoCov2 & 82.3 $\pm$  9.8 & 64.2 $\pm$ 12.0 & 81.2  $\pm$  6.1 &	77.7 $\pm$ 10.5 & 75.1 $\pm$ 5.8 & 77.8 $\pm$ 6.4 & 73.5 $\pm$ 10.4 & 58.2 $\pm$ 14.2 
   \\
   \hline
    \multicolumn{9}{c}{Models Trained on Surgical Videos} \\
    \hline
   MoCo v2~\cite{chen2020improved} & 85.7 $\pm$ 10.4 & 72.0 $\pm$ 9.1 & {\bf 85.4 $\pm$6.3}  & 84.9 $\pm$ 5.4 
   &
   78.4 $\pm$ 4.6 & 80.5  $\pm$ 5.1 & 78.5  $\pm$ 12.6 & 63.5  $\pm$ 13.6
   \\
   SwAV~\cite{caron2020unsupervised} & 84.7  $\pm$ 2.8 & 71.6 $\pm$ 7.9 & 84.9  $\pm$ 7.6  & 84.2 $\pm$ 6.3
   &
    77.3 $\pm$ 6.1 & 79.9  $\pm$ 6.3 & 77.8  $\pm$ 11.3 & 63.1  $\pm$ 11.5
   \\
SimSiam~\cite{chen2021exploring} & 83.9 $\pm$ 6.7 & 68.5 $\pm$ 5.5 & 84.5 $\pm$ 5.0 & 82.6 $\pm$ 5.7 &
 77.0 $\pm$ 5.3 & 79.1  $\pm$ 6.2 & 77.1  $\pm$ 8.2 & 62.8  $\pm$ 9.7
\\
RegionCL~\cite{xu2021regioncl} & 84.8 $\pm$ 11.4  &  69.8 $\pm$ 11.1 & 85.0 $\pm$  6.5 & 82.8 $\pm$ 10.9 
&
78.2  $\pm$ 10.4  &  79.9  $\pm$ 8.1 & 78.0  $\pm$ 9.5 & 63.4  $\pm$ 7.4  \\
  Ours & {\bf 87.3 $\pm$ 9.5} &  {\bf 73.5 $\pm$ 9.9} & { 85.1 $\pm$  7.6} & {\bf 86.1 $\pm$  6.9}
  &
  {\bf 81.1 $\pm$ 6.5} &  {\bf 80.7 $\pm$ 8.3} & {\bf 83.2 $\pm$  6.9} & {\bf 67.9 $\pm$  5.1}
  \\
\hline
\end{tabular}}
\end{table} 

To prove the effect of our proposed method, we compare our method with the state-of-the-art self-supervised learning approaches,~\emph{e.g.}, MoCo v2~\cite{chen2020improved}, SwAV~\cite{caron2020unsupervised}, SimSiam~\cite{chen2021exploring} and RegionCL~\cite{xu2021regioncl}. The results are shown in Table~\ref{tab:sota}.
Note that, our method uses MoCo v2~\cite{chen2020improved} as the self-supervised learning baseline.
%
We find that self-supervised training on ImageNet outperforms the model trained from scratch with a clear margin.
This indicates the powerful representation of the ImageNet pre-training, and motivates us to leverage semantics from it to improve the self-supervised training on surgical data.
It can be also found that self-supervised training on surgical videos outperforms that training on ImageNet~\cite{russakovsky2015imagenet}.
Combined with the distillation of semantics from ImageNet, our method can improve the baseline,~\emph{i.e.}, MoCo v2, over $1.6\%$ on Cholec80.

\subsection{Ablation Studies}
\begin{table}[t]
\centering
\caption{Ablation study on different transferring methods on Cholec80 dataset.
}\smallskip
\label{tab:distill}
\resizebox{0.8\columnwidth}{!}{
\begin{tabular}{c | c c c c }
%
\hline
   Method  & Accuracy & Precision & Recall & Jaccard \\
\hline
 Addition &  84.0 $\pm$ 10.5& 67.3 $\pm$ 11.1 &	82.7 $\pm$  8.7 &	81.6 $\pm$  8.3\\
 Concatenation  & 84.2 $\pm$  9.4 &	67.9 $\pm$ 11.7	& 83.1 $\pm$  8.6 &	81.0 $\pm$ 10.1\\
Initialization & 85.5 $\pm$ 10.9 &	 71.7 $\pm$ 8.1 &	83.8 $\pm$  7.2 &	83.2 $\pm$  8.4 \\
Distillation & {\bf 87.3 $\pm$ 9.5} &  {\bf 73.5 $\pm$ 9.9} &   {\bf 85.1 $\pm$  7.6} & {\bf 86.1 $\pm$  6.9} \\
\hline
\end{tabular}}
\end{table} 
\begin{figure}[t]%
\centering
    \includegraphics[width=1 \columnwidth]{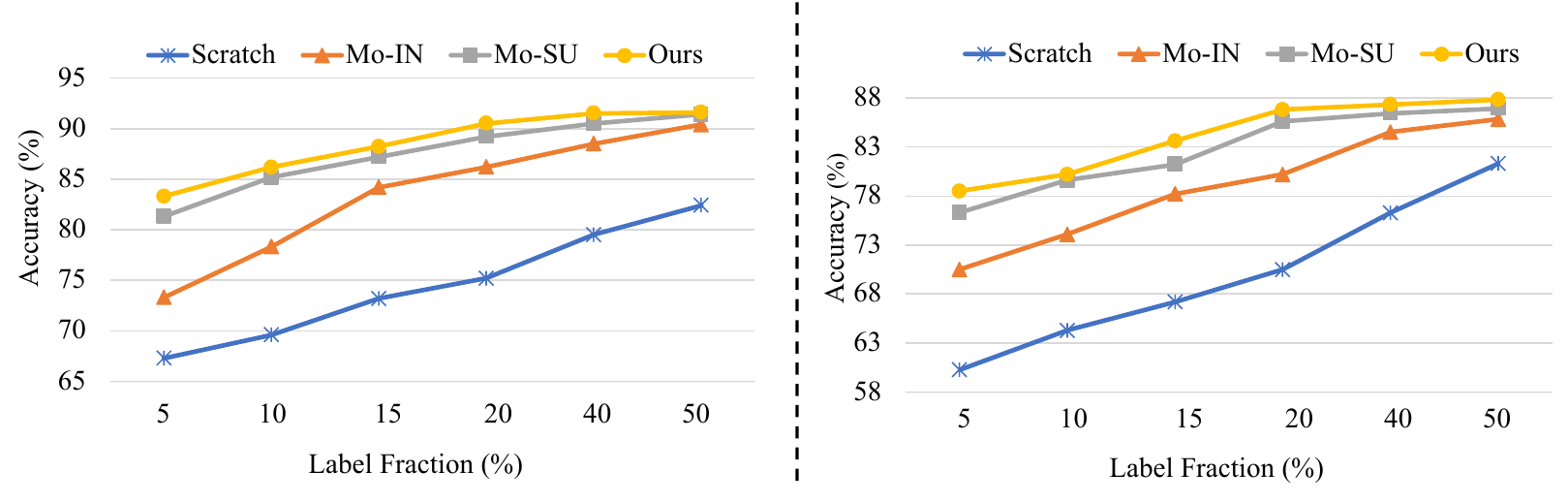}%
    \caption{ Accuracy for surgical phase recognition for different methods and varied sizes of label fractions on \textbf{Left}: Cholec80 and \textbf{Right}: MI2CAI16.
    ``Scratch" indicates training from scratch.``Mo-IN" and ``Mo-SU" refer to training from the model pre-trained on ImageNet and surgical data by MoCo v2, respectively.
    }
    \label{fig:label}%
\end{figure}
\begin{table}[t]
\centering
\caption{Ablation study on the semantic-preserving training on Cholec80 dataset. `Backbone' indicates training the backbone of the model, and `Head' means training the head projection of the model.
}\smallskip
\label{tab:semantic}
\resizebox{0.75\columnwidth}{!}{
\begin{tabular}{c c | c c c c }
%
\hline
   Backbone &  Head  & Accuracy & Precision & Recall & Jaccard \\
\hline
  \XSolidBrush & \XSolidBrush &  84.7 $\pm$ 11.0	& 67.5 $\pm$ 11.4& 84.2 $\pm$  8.2 & 79.6 $\pm$  8.8\\
  \XSolidBrush &\checkmark & {\bf 87.3 $\pm$ 9.5} &  {\bf 73.5 $\pm$ 9.9} & { 85.1 $\pm$  7.6} & {\bf 86.1 $\pm$  6.9}\\
  \checkmark & \XSolidBrush & 86.1 $\pm$ 11.8 &	 72.4 $\pm$ 9.4 &	{\bf 86.1 $\pm$  6.5}	& 84.8 $\pm$ 9.3\\
\checkmark & \checkmark & 85.2 $\pm$ 10.9&	 71.3 $\pm$ 10.4 &	85.1 $\pm$  5.7 &	85.8 $\pm$  6.5 \\
\hline
\end{tabular}}
\end{table} 

\vspace{1.5mm}
\noindent\textbf{Effect of distillation.}
Table~\ref{tab:distill} compares the performance of different transferring methods on Cholec80.
Specifically, we denote the feature maps generated from the teacher model and our model as $\mathbf{F}^t$ and $\mathbf{F}^s$ respectively.
`Addition' means the sum of $\mathbf{F}^t$ and $\mathbf{F}^s$,~\emph{i.e.}, $\mathbf{F}^t + \mathbf{F}^s $.
Similarly, `Concatenation' indicates the concatenation of 
$\mathbf{F}^t$ and $\mathbf{F}^s$.
`Initialization' means our model is initialized by the teacher model.
`Distillation' means using the distillation objective defined in Eq.~\ref{E:dis}.
It is clear that the distillation model is the best way to transfer the knowledge from the teacher model.

\vspace{1.5mm}
\noindent\textbf{Effect of the semantic-preserving training of free models.}
We conduct an ablation study on Cholec80 to evaluate the effect of the semantic-preserving training, and report the results in Table~\ref{tab:semantic}.
It is clear that not training the teacher model,~\emph{i.e.}, the first row in Table~\ref{tab:semantic}, cannot predict correct knowledge for surgical data, and would degrade the performance,~\emph{e.g.}, only achieve $84.7\%$ accuracy.
Furthermore, training too much would hurt the semantics learned from ImageNet; See the third and the last rows in Table~\ref{tab:semantic}.
We can find that our proposed semantic-preserving training (only updating parameters of the head projection (~\emph{i.e.}, $ h^t_q$ and $h^t_k$ ) and fixing the backbone (~\emph{i.e.}, $ f^t_q$ and $f^t_k$) can achieve the best performance.
Since trained with this way, the teacher model would not only maintain the semantics learned from ImageNet, but also can predict accurate knowledge for transferring.

\vspace{1.5mm}
\noindent\textbf{Label-efficiency analysis.}
To evaluate the label-efficiency of our self-training model, we fine-tune the pre-trained model on different fractions of labeled training data.
Note that in this ablation study, we train all parameters of the pre-trained model.
As shown in Fig.~\ref{fig:label}, our proposed approach achieves the best label-efficiency performance.
Furthermore, fine-tuning with fewer labels, the improvement gain is larger.

\section{Conclusion}

Unlike existing self-supervised papers that design new pretext tasks, this paper proposes a novel free lunch method for self-supervised learning with surgical videos by distilling knowledge from free available models. 
Our motivation is that the publicly released model trained on natural images shows comparable performance with the model trained on the surgical data.
To leverage the rich semantics from the available models, we propose a semantic-persevering method to train a teacher model that contains prior information and can produce accurate predictions for surgical data.
Finally, we introduce a distillation objective to the self-training on surgical data to enable the model to learn extra knowledge from the teacher model.
Experimental results indicate that our method can improve the performance of the existing self-supervised learning methods.

\section{Acknowledgement}
This work was supported by a research grant from HKUST-BICI Exploratory Fund (HCIC-004) and a research grant from Shenzhen Municipal Central Government Guides Local Science and Technology Development Special Funded Projects (2021Szvup139), and under the RIE2020 Industry Alignment Fund – Industry Collaboration Projects (IAF-ICP) Funding Initiative, as well as cash and in-kind contribution from the industry partner(s).
\bibliographystyle{splncs04}
\bibliography{miccai_bib}
%




\end{document}